\documentclass[letterpaper]{article} 
\usepackage{aaai24}  
\usepackage{times}  
\usepackage{helvet}  
\usepackage{courier}  
\usepackage[hyphens]{url}  
\usepackage{booktabs}
\usepackage{graphicx} 
\usepackage{natbib}  
\usepackage{caption} 
\usepackage{algorithm}
\usepackage{algorithmic}

\usepackage{newfloat}
\usepackage{listings}
\DeclareCaptionStyle{ruled}{labelfont=normalfont,labelsep=colon,strut=off} 
\floatstyle{ruled}
\newfloat{listing}{tb}{lst}{}
\floatname{listing}{Listing}
\title{Relation Extraction from News Articles (RENA): A Tool for Epidemic Surveillance}
\author{
    Jaeff Hong\equalcontrib\textsuperscript{\rm 2}, Duong Dung\equalcontrib\textsuperscript{\rm 1}, Danielle Hutchinson\textsuperscript{\rm 1}, Zubair Akhtar\textsuperscript{\rm 1}, Rosalie Chen\textsuperscript{\rm 1}, Rebecca Dawson\textsuperscript{\rm 1}, Aditya Joshi\footnote{Corresponding author}\textsuperscript{\rm 2}, Samsung Lim\textsuperscript{\rm 1 \rm 3}, C Raina MacIntyre\textsuperscript{\rm 1} and Deepti Gurdasani\textsuperscript{\rm 1 \rm 4 \rm 5}, \\
}
\affiliations{
    \textsuperscript{\rm 1}Kirby Institute, University of New South Wales, Sydney, Australia\\
 \textsuperscript{\rm 2}School of Computer Science \& Engineering, University of New South Wales, Sydney, Australia\\
\textsuperscript{\rm 3}School of Civil and Environmental Engineering, University of New South Wales, Sydney, Australia\\ 
 \textsuperscript{\rm 4}William Harvey Research Institute, Queen Mary University of London, London, UK\\
 \textsuperscript{\rm 5}School of Medicine, University of Western Australia, WA, Australia\\

 jaeff.hong@gmail.com, \{duong.dung, danielle.hutchinson, zubair.akhtar, rosalie.chen, rebecca.dawson, aditya.joshi, s.lim, r.macintyre, d.gurdasani\}@unsw.edu.au

}

\usepackage{bibentry}

\begin{document}

\maketitle

\begin{abstract}
Relation Extraction from News Articles (RENA) is a browser-based tool designed to extract key entities and their semantic relationships in English language news articles related to infectious diseases. Constructed using the React framework, this system presents users with an elegant and user-friendly interface. It enables users to input a news article and select from a choice of two models to generate a comprehensive list of relations within the provided text. As a result, RENA allows real-time parsing of news articles to extract key information for epidemic surveillance, contributing to EPIWATCH, an open-source intelligence-based epidemic warning system.
\end{abstract}

\section{Introduction}
Online news websites are a valuable source of information about real-world events with several potential applications. One such application is the use of news articles for text-based epidemic intelligence~\cite{joshi2019survey}. EPIWATCH is an open-source intelligence-based early warning system for epidemics that parses vast amounts of data in real-time to build a structured repository of data to study changing trends in diseases and syndromes, geographically, over time~\cite{macintyre2023artificial}. A relevance classifier selects news articles from multiple sources~\cite{?}, which are then reviewed by expert analysts daily to generate structured data such as name of disease, number of cases and so on. This structured information allows identification of potential outbreaks earlier than traditional surveillance systems, alerting health authorities, thereby allowing for quicker outbreak response, and preventing further spread, ultimately saving lives~\cite{puca2020using}.

The manual task of extracting structured information from news articles maps to the natural language processing (NLP) task of relation extraction~\cite{banko2008tradeoffs, kumar2017survey}. Our browser-based tool, \textit{Relation Extraction from News Articles (RENA)}, uses decoder-only foundation models (also known as `large language models', \textit{i.e.}, LLMs) to extract semantic relations in infectious disease-related news articles in the English language\footnote{With appropriate choice of foundation models, RENA can be extended to languages other than English. We are interested in doing so.} at the document level. When an epidemiologist enters a news article, RENA extracts a list of entities and relations present in the article. This streamlines an automated process for epidemiologists and researchers who need a method to acquire a large congregate of structured relations from their selected article. By making use of RENA, they can be aided in their research without having to read a large set of news articles or delve into manual data curation.

Whilst many previous relation extraction (RE) tasks have focused on the sentence level, it is evident that many relations exist between different sentences, presenting another challenge in extracting relations at the document level~\cite{xu2021document}. We assume a simplistic definition of a relation: a relation connects exactly two entities. In the context of epidemic intelligence, the sentence `A patient died due to COVID-19 today' results in the relations \{death number: '1', relation: 'death of', infectious disease: COVID-19'\}, \{death number: '1', relation: 'occurred on', event date: 'today'\} and \{infectious disease: COVID-19, relation: 'occurred on', event date; 'today'\}. In this specific case, if two relations are identified, the third can be inferred. 

RENA can be used by epidemiologists, public health officials and teachers or students of public health to extract information from news articles of interest. While the utility of RENA is for infectious disease-related epidemic intelligence, it can potentially be used for news articles across domains and application areas, including journalists/news publishers who want to verify and investigate information across multiple sources.
\begin{figure*}[h]
\centering
\includegraphics[width=1.0\textwidth, height=0.25\textheight]{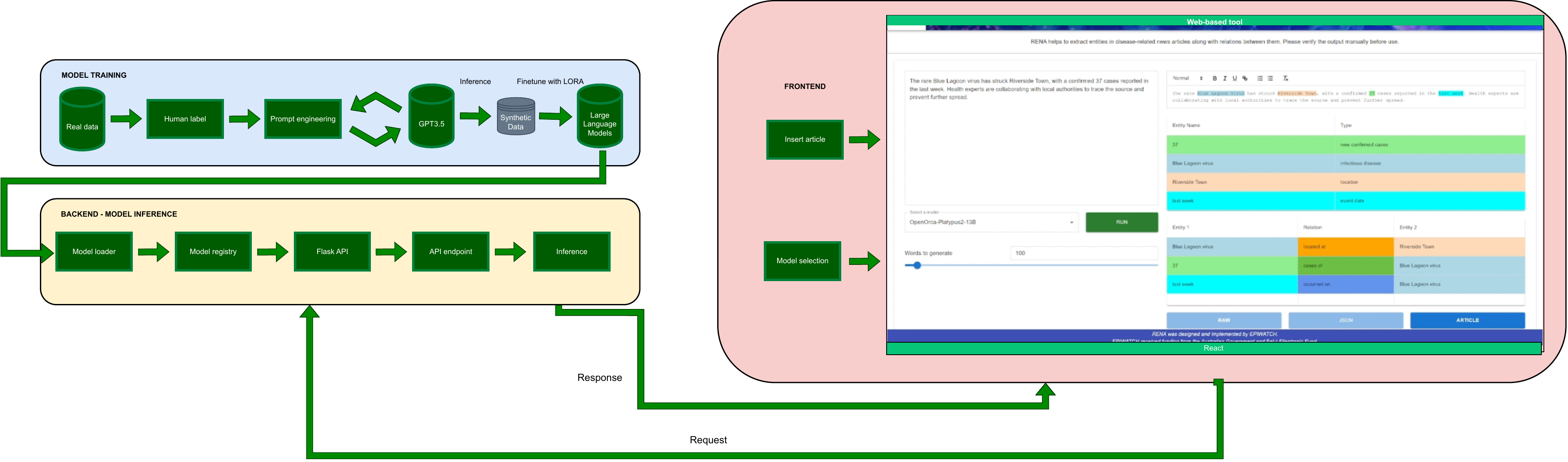}
\caption{Architecture}
\label{fig:architecture}
\end{figure*}

\section{Architecture}
Figure~\ref{fig:architecture} shows the architecture of RENA. In the model training phase, we generated a set of 300 synthetic articles, annotated with relevant entities and relationships using 1-shot prompting with OpenAI API~\footnote{?; Accessed on 3rd October, 2023.} (gpt-3.5-turbo-16k) (Details in Supplementary Document.). The generated output (articles, and labeled entities and relations) were used for fine-tuning two large language models including \texttt{OpenOrca-Platypus2-13B}\footnote{\url{https://huggingface.co/Open-Orca/OpenOrca-Platypus2-13B}; Accessed on 3rd October, 2023.} and \texttt{Mythical-Destroyer-V2-L2-13B}\footnote{\url{https://huggingface.co/Sao10K/Mythical-Destroyer-V2-L2-13B}; Accessed on 3rd October, 2023.}, both based on Meta's LLaMa2 13B model~\cite{touvron2023llama}, using quantized low-rank adaptation (QLoRa)~\cite{dettmers2023qlora}. We run the training process for three epochs. In the inference phase, RENA allows for the selection of the two different models specified above. Our backend is built with Python Flask, where the model is both loaded and executed for inference. The output is then sent to our frontend, developed in React, to display the response.

We evaluated the two models included in RENA on 10 randomly selected articles in English, selected from the EPIWATCH database, an existing database with more than 1 million news articles relating to infectious disease outbreaks searched through online news, and prioritised using machine learning-based classification supplemented with manual curation by experts. We computed precision, recall and F1 score, comparing the predicted to labeled output on the ten articles for each model. We assessed the accuracy for named entity recognition (NER) and RE separately. RE was only evaluated for relevant entities that were recognised, so both metrics should be considered when assessing the models. Table~\ref{tab1} shows the precision, recall and F-score values for the models included in RENA.



\begin{table}[]
\begin{tabular}{lllll}
\toprule
 Model  & Eval & Precision &  Recall  &  F1  \\ \midrule
 Open-Orca-13b &  NER & 0.93 & 0.66 & 0.77 \\
 Open-Orca-13b  & RE  & 0.88 & 0.88 & 0.88 \\
 Mythical-Destroyer-13b  & NER & 0.96  & 0.57 & 0.71 \\
 Mythical-Destroyer-13b & RE & 0.97 & 0.97 & 0.97 \\ \bottomrule
\end{tabular}
\caption{Evaluation of models in RENA.}
\label{tab1}
\end{table} 


\section{User Interface}

Figure 1 is a snapshot of the web interface hosting RENA. On the webpage includes:
\begin{itemize}
    \item \textbf{Input:} The input contains the text field that the user can input their unstructured article into, from which the model will extract entities and relations.
    \item \textbf{Select Model:} The select model box contains a dropdown to select between the 2 models to be used for generation.
    \item \textbf{Submit:} Begins the generation process.
    \item \textbf{Max tokens:} Max tokens is a slider to determine how many tokens of output will be produced by the model.
    \item \textbf{Output:} The model will generate the output (relations) into this box.
    \item \textbf{Raw Button:} The raw generated output of the model will be displayed in the output box when clicked.
    \item \textbf{JSON Button:} The output will be converted into a JSON format and be displayed in the output box when clicked.
    \item \textbf{Article Button:} The article will be displayed in the output box and the entities will be highlighted in the text and color coded. A table showing all the entity and entity types as well as another table showing the relations will also be shown. 
\end{itemize}

\section{Summary \& Future Work}
RENA is a tool that uses QLoRA to finetune LLMs and extract disease-related relations from news articles. It offers a way to automate the creation of structured data from an inputted unstructured news article and output relatively accurate results. RENA will enable users interested in researching diseases to parse large amounts of data to potentially aid in the surveillance and prevention of diseases. We plan to extend RENA to multilingual news articles. In addition, different relations can help delineate different disease events, allowing detection of independent outbreaks involving different infections or locations from news articles.  
\bibliography{aaai24}

\begin{thebibliography}{8}
\providecommand{\natexlab}[1]{#1}

\bibitem[{Banko and Etzioni(2008)}]{banko2008tradeoffs}
Banko, M.; and Etzioni, O. 2008.
\newblock The tradeoffs between open and traditional relation extraction.
\newblock In \emph{Proceedings of ACL-08: HLT}, 28--36.

\bibitem[{Dettmers et~al.(2023)Dettmers, Pagnoni, Holtzman, and Zettlemoyer}]{dettmers2023qlora}
Dettmers, T.; Pagnoni, A.; Holtzman, A.; and Zettlemoyer, L. 2023.
\newblock Qlora: Efficient finetuning of quantized llms.
\newblock \emph{arXiv preprint arXiv:2305.14314}.

\bibitem[{Joshi et~al.(2019)Joshi, Karimi, Sparks, Paris, and Macintyre}]{joshi2019survey}
Joshi, A.; Karimi, S.; Sparks, R.; Paris, C.; and Macintyre, C.~R. 2019.
\newblock Survey of text-based epidemic intelligence: A computational linguistics perspective.
\newblock \emph{ACM Computing Surveys (CSUR)}, 52(6): 1--19.

\bibitem[{Kumar(2017)}]{kumar2017survey}
Kumar, S. 2017.
\newblock A survey of deep learning methods for relation extraction.
\newblock \emph{arXiv preprint arXiv:1705.03645}.

\bibitem[{MacIntyre et~al.(2023)MacIntyre, Chen, Kunasekaran, Quigley, Lim, Stone, Paik, Yao, Heslop, Wei et~al.}]{macintyre2023artificial}
MacIntyre, C.~R.; Chen, X.; Kunasekaran, M.; Quigley, A.; Lim, S.; Stone, H.; Paik, H.-y.; Yao, L.; Heslop, D.; Wei, W.; et~al. 2023.
\newblock Artificial intelligence in public health: the potential of epidemic early warning systems.
\newblock \emph{Journal of International Medical Research}, 51(3): 03000605231159335.

\bibitem[{Puca and Trent(2020)}]{puca2020using}
Puca, C.; and Trent, M. 2020.
\newblock Using the Surveillance Tool EpiWATCH to Rapidly Detect Global Mumps Outbreaks.
\newblock \emph{Global Biosecurity}, 2(1).

\bibitem[{Touvron et~al.(2023)Touvron, Lavril, Izacard, Martinet, Lachaux, Lacroix, Rozi{\`e}re, Goyal, Hambro, Azhar et~al.}]{touvron2023llama}
Touvron, H.; Lavril, T.; Izacard, G.; Martinet, X.; Lachaux, M.-A.; Lacroix, T.; Rozi{\`e}re, B.; Goyal, N.; Hambro, E.; Azhar, F.; et~al. 2023.
\newblock Llama: Open and efficient foundation language models.
\newblock \emph{arXiv preprint arXiv:2302.13971}.

\bibitem[{Xu, Chen, and Zhao(2021)}]{xu2021document}
Xu, W.; Chen, K.; and Zhao, T. 2021.
\newblock Document-level relation extraction with reconstruction.
\newblock In \emph{Proceedings of the AAAI Conference on Artificial Intelligence}, volume~35, 14167--14175.

\end{thebibliography}

\end{document}


\maketitle

\section{Prompt used to generate synthetic data}

We used the following prompt in GPT3.5 to generate synthetic data:

\textbf{System instruction:} 
\begin{lstlisting}
You are an AI content creator who helps to write news about epidemic around the world.
\end{lstlisting}

\textbf{User instruction}:  
\begin{lstlisting}
I have an example articles:
Example:
Title: Ebola Virus Epidemic: 25 new cases reported in Democratic Republic of Congo
Publication date: October 5, 2021

Content: The Ebola virus epidemic in the Democratic Republic of Congo has shown a recent surge in new cases. According to the World Health Organization, 25 new cases were reported in the past week, bringing the total number of cases to 3,825, with 2,589 deaths. The outbreak, which began in August 2018, is the second deadliest in history, after the 2014-2016 West Africa epidemic that killed more than 11,000 people. The WHO and its partners are continuing to work on the ground to control the spread of the virus, including through vaccination campaigns and community engagement. However, the ongoing conflict in the region and resistance from some communities have made the response more challenging.

Please help me to create another one. 
\end{lstlisting}

\section{Prompt used to generate entity and relation triplets for synthetic data}

In designing our prompt to identify entities and relations we aimed to be as specific and simple as possible. The prompt engineering was tailored towards striking a balance between specificity and simplicity where we found by including the list of entities of relations in the prompt, the models could better understand the task at hand and generate responses that aligned closely with expectations. We used the following prompt in GPT3.5 to extract entities and relationships within synthetic data generated using GPT3.5:

\textbf{System instruction:} 
\begin{lstlisting}
You are a smart and intelligent Relation Extraction (RE) system for diseases information.
\end{lstlisting}


\textbf{User instruction}:  

\begin{lstlisting}
  I have the list of relations including:
- "located at" is between: disease - location, symptom/syndrome - location, case number - location
- "occurred on" is between: disease - date, symptom/syndrome - date, case number - date
- "are symptoms of" is between: disease - symptom/syndrome
- "deaths of" is between: disease - death numbers
- "cases of" is between: disease - case numbers, symptom/syndrome - case numbers
- "caused by" is between: disease - pathogen
- "affected by" is between: people - disease
Extract all relations between infectious disease, pathogen, symptoms, syndromes, case numbers, date and location from the main body of the article.                   
Example:
Article: Laos reports two H5N1 avian influenza poultry outbreaks, WHO follow-up on human case - Outbreak News Today Follow Privacy Policy US News Europe Asia Africa Latin America and the Caribbean Canada Indian subcontinent Australia Middle East BREAKING Poland reports 12 trichinosis in first half of 2023 Virginia Tech researchers to study lingering Lyme disease symptoms, Awarded NIH grant Malawi cholera update Denmark reports increase in pertussis cases in May and June 2023 Dominican Republic: 11 cases of acute diarrhea, suspected of cholera Outbreak News Radio Interviews Drugs and antibiotics Vaccines Research Blogs TRENDING US News Latin America and the Caribbean Animal diseases Home Asia Laos reports two H5N1 avian influenza poultry outbreaks, WHO follow-up on human case Laos reports two H5N1 avian influenza poultry outbreaks, WHO follow-up on human case by News Desk November 19, 2020 Asia, Headlines No Comments By NewsDesk @ bactiman63 Poultry outbreaks Last Saturday, Dr Vilayphone Vorraphim, Director General of the Department of Livestock and Fisheries with the Ministry of Agriculture and Forestry in Vientiane, Laos reported two Highly pathogenic avian influenza (HPAI) virus (H5N1) outbreaks in two villages in Saravane province. The report to the the World Organisation for Animal Health (OIE) noted that the two outbreaks occurred in the villages of Khantharath and Nakhoysao. More than 500 birds died in the two villages due to the virus. The case is a one-year-old female who developed symptoms of fever, productive cough, difficulty breathing and runny nose on 13 October 2020. She was hospitalized for her illness on 16 October and discharged on 19 October. As part of severe acute respiratory infection (SARI) sentinel surveillance, a specimen was collected on the date of hospitalization and confirmed to be positive for avian influenza A (H5N1) on 28 October by reverse transcription polymerase chain reaction (RT-PCR) at the National Centre for Laboratory and Epidemiology (NCLE).
\end{lstlisting}

\textbf{Output:}
\begin{lstlisting}
1) "infectious disease": "avian influenza (HPAI) virus (H5N1)", "relation": "located at", "location": "Saravane province"
2) "infectious disease": "avian influenza (HPAI) virus (H5N1)", "relation": "located at", "location": "Khantharath"
3) "syndrome": "fever, productive cough, difficulty breathing and runny nose", "relation": "occurred on", "event date": "13 October 2020"
4) "overall confirmed deaths": "500", "relation": "deaths of", "host": "one-year-old female"
5) "new confirmed cases": "two", "relation": "cases of", "infectious disease": "avian influenza (HPAI) virus (H5N1)"
Article: {content}
Output:
"""
\end{lstlisting}

\section{Configuration Hyperparameters}

\begin{itemize}
    \item model\_type: LlamaForCausalLM
    \item tokenizer\_type: LlamaTokenizer
    \item is\_llama\_derived\_model: true
    \item load\_in\_8bit: false
    \item load\_in\_4bit: true
    \item strict: false
    \item val\_set\_size: 0.01
    \item adapter: qlora
    \item sequence\_len: 4096
    \item sample\_packing: true
    \item pad\_to\_sequence\_len: true
    \item lora\_r: 64
    \item lora\_alpha: 32
    \item lora\_dropout: 0.05
    \item lora\_target\_linear: true
    \item gradient\_accumulation\_steps: 4
    \item micro\_batch\_size: 1
    \item num\_epochs: 3
    \item optimizer: paged\_adamw\_32bit
    \item  lr\_scheduler: cosine
    \item learning\_rate: 0.0002
    \item train\_on\_inputs: false
    \item group\_by\_length: false
    \item bf16: false
    \item fp16: true
    \item tf32: false
    \item gradient\_checkpointing: true
    \item logging\_steps: 1
    \item flash\_attention: false
    \item warmup\_steps: 10
    \item eval\_steps: 20
    \item weight\_decay: 0.0
    \item special\_tokens:
      \item bos\_token: "$<s>$"
      \item eos\_token: "$</s>$"
      \item unk\_token: "$<unk>$"
\end{itemize}

\section{Prompt used for evaluation of fine-tuned models}


\textbf{Prompt template as follows:}
\begin{lstlisting}
### Instruction:  You are an AI assistant. You will be given a task. You must generate a detailed and long answer.
 
            I have the list of relations including: 
            - "located at" is between: disease - location, symptom/syndrome - location, case number - location
            - "occurred on" is between: disease - date, symptom/syndrome - date, case number - date
            - "are symptoms of" is between: disease - symptom/syndrome
            - "deaths of" is between: disease - death numbers
            - "cases of" is between: disease - case numbers, symptom/syndrome - case numbers
            - "caused by" is between: disease - pathogen
            - "affected by" is between: people - disease

            Extract all relations between infectious disease, pathogen, symptoms, syndromes, case numbers, date and location from the main body of the article.
            Article: India has reported a surge in H1N1 influenza cases in the past week, with more than 3,000 new cases and 27 deaths. The majority of cases have been reported in the states of Maharashtra, Karnataka, and Telangana. Health officials have urged people to take precautions, including getting vaccinated, washing hands frequently, and avoiding crowded places. The H1N1 virus, also known as swine flu, first emerged in 2009 and caused a global pandemic. While it is now considered a seasonal flu, it can still cause severe illness and death, particularly in vulnerable populations such as young children, pregnant women, and the elderly.
### Response:
\end{lstlisting}